# Hybridization of Capsule and LSTM Networks for unsupervised anomaly detection on multivariate data

Ayman Elhalwagy and Tatiana Kalganova, *Member, IEEE*

*Abstract*— **Deep learning techniques have recently shown promise in the field of anomaly detection, providing a flexible and effective method of modelling systems in comparison to traditional statistical modelling and signal processing-based methods. However, there are a few well publicised issues Neural Networks (NN)s face such as generalisation ability, requiring large volumes of labelled data to be able to train effectively and understanding spatial context in data. This paper introduces a novel NN architecture which hybridises the Long-Short-Term-Memory (LSTM) and Capsule Networks into a single network in a branched input Autoencoder architecture for use on multivariate time series data. The proposed method uses an unsupervised learning technique to overcome the issues with finding large volumes of labelled training data. Experimental results show that without hyperparameter optimisation, using Capsules significantly reduces overfitting and improves the training efficiency. Additionally, results also show that the branched input models can learn multivariate data more consistently with or without Capsules in comparison to the non-branched input models. The proposed model architecture was also tested on an open-source benchmark, where it achieved state-of-the-art performance in outlier detection, and overall performs best over the metrics tested in comparison to current state-of-the art methods.**

## I. INTRODUCTION

Time Series data analysis is a prominent field of research due to the significant demand stemming from increasingly larger datasets being acquired in industrial and commercial environments. The automation of this analysis has been integral to the advancement of Industry 4.0 [1], which refers to the automation of industrial processes. One important use case for time series analysis is outlier detection, which is an important part of the function of intelligent systems in the context of fault diagnosis.

There are various traditional approaches to fault detection by way of hardware redundancy in literature [2]. However, a rising need in industry for a lightweight and cost-effective solution to fault detection as a result of Industry 4.0 has encouraged the development of soft sensing systems, which use the existing sensors in a system to infer further information regarding the system. Recently, Neural Networks (NNs) have been identified as an effective tool in data analysis and fault detection due to their unique ability to be trained to identify numerical relationships in different forms of data [3]. They provide an advantage over traditional signal processing and statistical techniques due to the level of complexity that they can model the data, as well as being generalizable to similar types of data [3]. Furthermore, there is minimal data manipulation needed for the use of NNs, which can simplify the implementation of such systems.

For time series analysis, Recurrent Neural Network (RNN) based models such as the Long-Short Term Memory (LSTM) network [4] are generally used due to their ability to identify dependencies in sequential data using their internal memory [5], [6], however some recent works have utilised the Convolutional Neural Network (CNN), which has proved to be a powerful tool in image classification [7], [8], in time series forecasting [9] and outlier detection tasks [10], [11]. Furthermore, the hybridisation of the aforementioned layers has also been shown to be a method of improving outlier detection performance [12], [13]. However, it is well documented that the CNN has a fundamental flaw in understanding spatial context in data; this is most prominently demonstrated in image classification tasks. The Capsule Network (CapsNet) [14] was proposed by Hinton et al to address this flaw and has successfully shown state-of-the-art performance in image classification tasks with different variants of the network [15]. Additionally, some work has been done utilising the CapsNet for use on time series data in its raw format [16], [17], but predominantly using image representations[18], [19]; most approaches use an image representation of the data as the CapsNet has been proven to improve performance in this context. In this paper, we propose the hybridisation of the CapsNet and the LSTM Network in a branched Autoencoder architecture for use on raw time series

This paragraph of the first footnote will contain the date on which you submitted your paper for review, which is populated by IEEE. It is IEEE style to display support information, including sponsor and financial support acknowledgment, here and not in an acknowledgment section at the end of the article. For example, "This work was supported in part by the U.S. Department of Commerce under Grant BS123456." The name of the corresponding author appears after the financial information, e.g. *(Corresponding author: M. Smith).* Here you may also indicate if authors contributed equally or if there are co-first authors.

The next few paragraphs should contain the authors' current affiliations, including current address and e-mail. For example, First A. Author is with the National Institute of Standards and Technology, Boulder, CO 80305 USA (e-mail: author@ boulder.nist.gov).

Second B. Author, Jr., was with Rice University, Houston, TX 77005 USA. He is now with the Department of Physics, Colorado State University, Fort Collins, CO 80523 USA (e-mail: author@lamar.colostate.edu).

Third C. Author is with the Electrical Engineering Department, University of Colorado, Boulder, CO 80309 USA, on leave from the National Research Institute for Metals, Tsukuba 305-0047, Japan (e-mail: author@nrim.go.jp).

Mentions of supplemental materials and animal/human rights statements can be included here.

Color versions of one or more of the figures in this article are available online at http://ieeexplore.ieee.org

data.

The contributions of this paper are summarised as follows:

1) A hybridisation of the LSTM layer and the Capsule layer is implemented in a novel branched input, merged output model architecture for use on raw multivariate time series data
2) The model is tested on a real-world dataset and benchmarked on another real-world dataset against the state-of-the-art anomaly detection methods in the field for a performance comparison

## II. MOTIVATION AND RELATED PUBLICATIONS

This section will first explore the advantages of soft sensing methods over traditional hardware redundancy techniques for fault detection, then outline the benefits of NN soft sensing methods over traditional system modelling and signal processing techniques. NN based fault detection methods will then be reviewed so that a justification for the proposed method can be made.

Fault detection systems have been researched and improved extensively over the last two decades due to the intense demand to automate industrial processes, also known as Industry 4.0 [20]. There have been numerous approaches that aim to be effective in detecting different types of faults in different systems, due to the nature of the usage of the system or other reasons relating to the susceptibility of the system to certain faults. Some approaches for fault detection have involved using methods and techniques such as redundancy for sensors, sometimes paired with analytical redundancy methods[21], [22].

The common issue with these proposed solutions is that they involve the installation and maintenance of physical hardware to monitor the sensors or the system, which naturally means that they may require redundancy in more sensitive use cases: for example, that require the monitoring of life-threatening substances with very sensitive sensors. Furthermore, this guarantees an increase in the operating costs of these solutions due to increased energy usage and maintenance costs and provides another barrier to the goal of achieving automation. However more recently, soft sensing methods have been explored with the goal of using the information available from the sensors already implemented in the system to calculate an estimate of the quality of data being collected. This approach provides an economical and cost-effective alternative to physical systems by not needing to implement any additional physical hardware that could be expensive to buy or maintain whilst achieving robust fault detection scores that are comparable to and even better than physical systems.

### A. Soft Sensing Methods for Fault Detection

Statistical analysis and signal processing are frequently used methods in the field of anomaly detection. As a method of soft sensing, they are able to overcome the drawbacks of physical hardware monitoring and provide a robust method of data inference. For instance in [23] a dynamic model is proposed that is able to utilise the existing supervisory control and data acquisition (SCADA) system in wind turbines to dynamically model the relationship between the sensor readings by a parameter estimation process for the purpose of fault detection. A frequency domain analysis is used to determine damage sensitive indices which are then compared to the model sensor. The technique is tested on a 5-year wind turbine dataset where the system was able to detect faults as well as perform fault prognosis. Whilst the method is clearly effective in the specified use case, the flexibility of the method for other use cases comes into question as in-depth knowledge about the system and the relationships between the variables being analysed was utilised to be able to create the initial model. This issue is also mentioned in [24] where the authors concluded from their survey of outlier detection techniques that model-driven methods are heavily dependent on the understanding of the data being analysed. The lack of flexibility of such techniques is also mentioned, due to the heavy tailoring that must be made to the models for each dataset. This is a trend across many signal processing techniques including for motor condition monitoring where [25] noted in their state-of-the review of outlier detection techniques the lack of flexibility of data analysis techniques such as acoustic analysis and motor current signal analysis (MCSA) in detecting a wide range of faults that could occur within the system.

More recently, Machine Learning (ML) has been heavily utilised in literature for the modelling of such systems. As well as being a soft sensing technique, ML is able to provide a higher degree of flexibility in terms of application as well as being generally easier to implement than the aforementioned techniques. Various examples of literature can be found that utilise proposed NN models in multiple use cases and datasets [6], [26]. However, this is not to say that generalisation is still not an issue with NNs. The main issue found in literature with NNs is the importance of data volume and representation in being able to train NNs effectively. Most NN types such as the CNN and the LSTM require large quantities of data to effectively learn the shape and features of the data, and some methods even require the labelling of the data before training, known as supervised learning [12], which is very time-consuming and costly as this is usually a manual process. Furthermore, with some types of data it is very difficult to distinguish faults and anomalies in raw sequential format.

To address these discussed issues with NNs, researchers have opted to combine signal processing techniques with NNs where applicable in order to utilise the advantages provided by the former with data representation and the latter in flexibility and ease of use. This approach has seen great success in motor fault detection [27], [28], where in these cases frequency domain transformations were used to enhance the representation of the data for use with LSTM networks. Additionally, the use of various types of CapsNets in numerous cases was found to improve training and classification performance over smaller datasets [29], [30]. As well as hybridising signal processing and ML, many literatures also propose the hybridisation of NN types to take advantage of their advantages with different types of data; one popular hybridisation for TS data is RNNs and

CNNs [12].

### B. LSTM based Autoencoder

The LSTM network [4] has lately grown in popularity due to the overwhelming demand for time series analysis and forecasting in commercial environments, hence fuelling the demand for more research into the improvement of its performance. One recently proposed method [31] explored the usage of the LSTM layer in an Autoencoder architecture. The authors correctly identified that a large number of the current machine learning methods that are used are unsuitable for use practically, as they usually require the use of labelled data which is impractical with time series data due to the large volumes being constantly produced. Furthermore, the authors go on to evaluate classical anomaly detection methods such as Support Vector Machines and Isolation forests as being flawed since they fail to account for the temporal aspect of the time series data and only take the current data into account. In addition to this, they demonstrate, with an initial experiment, that other methods such as signal decomposition are only effective when used with periodic data, so their use is limited in that aspect. However, the authors [31] noted that the simplicity of this method is an advantage over the LSTM autoencoder approach that was being explored, but the necessity of manual parameter selection was considered a drawback.

The proposed method in [31] uses the sliding window algorithm to feed the data into the LSTM Autoencoder, for which the number of layers and LSTM cells are optimised. The neural network was trained by fitting the output to the input signal, and the mean absolute error between the prediction and signal was used as the threshold for testing for anomalies. The system was tested on sound files from the DCASE dataset which were down sampled to 16000 samples per second, and the sliding window would take 1-second steps. Results [31] show an 87% accuracy for anomaly detections, with the correct location identified 91.7% of the time.

Evaluating the approach used in this paper [31], it shows promise with the accuracy of detection and the wide application of its usage, but various drawbacks were identified: the authors selectively used data that was loud enough to be detected by the autoencoder and did not explore the sensitivity of detection. Furthermore, the authors assumed that the training data acquired was "clean" of any anomalies, which could be a reasonable assumption to make since the data was taken from an established dataset, but in a real-life use case, this may not be the case. However, since this is an unsupervised approach, it is expected that all initial errors will not be identified unless extensive data analysis is carried out before training the system, or if previous knowledge about the operation of the system being analysed is acquired. Furthermore, due to the sliding window approach, the location of the anomaly was a parameter that had to be measured, which could be overcome if a different approach to parsing the data was used. One main issue that the LSTM faces is its overfitting when used with gradient descent learning optimisation algorithms. Although very careful tweaking of hyperparameters can help to reduce this issue, this is often highly inefficient and time consuming and with complex datasets is sometimes unavoidable.

### C. Capsule Network application in time series data analysis

The CapsNet [14] is a novel neural network developed to overcome issues faced with the spatial context of image classification that is encountered when using prevalent image classifications such as convolutional neural networks. It does this by "encapsulating" the entity being described in a vector format, where the length describes the probability of existence and the orientation of the vector describes the entity's characteristics, such as orientation and the special context that other traditional neural networks cannot capture. The original idea was conceptualized in [14] by Hinton et al, but some literatures [32] have built on this work by adjusting the architecture to work with Adaptive Gradient Descent optimisation algorithms for image classification. However, the effectiveness of the latter approach has not been fully explored for raw time series data at the time of writing. This architecture is mainly used for image classification as demonstrated by [33] for brain tumour classification using MRI images and [34] for "Hyperspectral Image Classification", but more recently its usage has been explored limitedly in a time series use case.

A few papers currently exist that utilise this neural network architecture for time series tasks [35]. However, the approaches that most papers use is to transform the data into an image representation, which has already been identified as a powerful usage of this network. For example, one proposed approach [36] aimed to utilise capsule networks to address an issue with the detection of short circuit faults in a power network transmission line. The authors state that the raw signals are difficult to analyse for this task and analyse methods for time series feature extraction in the literature review. The Fourier transform was identified as a prevalent method of frequency domain analysis, however, the authors identify that the former does not take the temporal context into account which, for the use case that this paper covers, is an essential factor. Therefore, another paper was outlined [37] which overcame this issue using the discrete wavelet transform, which is able to provide information from both the frequency and time domain. The authors [36] note that these signal processing techniques require a high level of expert knowledge in order to leverage properly to produce good results, and image representation techniques can extract more significant features more efficiently in comparison to signal processing methods. Using this information, the authors propose a deep learning approach that overcomes the issues that current machine learning methods have with poorer feature representation due to scalar values being used and max pooling inhibiting the information learned by the neural network.

A 3-phase voltage system is first simulated [36] so that different fault models can be identified and simulated, and the signal processing can be applied and tested for robustness. The discrete wavelet transform that was identified as a superior method to the Fourier analysis is used in combination with a high pass and low pass filter in order to filter the signal noise, and a polar representation of the signal is acquired which is then represented as a Gramian Angular Field (GAF), proposed in [38], which is a method of time series image representation. The authors reasoning for this is a more feature-rich representation of the data in comparison to other image representations, due to the preservation of the temporal context as well as the other points identified in their literature review.

The proposed approach in [36] utilises the capsule network to overcome the discussed issues in the literature review. Furthermore, the architecture proposed uses convolutional layers that accept 6 inputs corresponding to the number of signals in the 3-phase voltage system transformed into pictures. The convolutional operation then outputs to a self-attention layer, which the author [36] claims produces promising results by focusing on the more relevant areas in the GAF image. A Rectified Linear Unit (Relu) activation is then used before 2 capsule layers which output the classification results. The neural network uses a novel technique referred to as weight sharing which connects the neurons in a different configuration to a normal fully connected network so that the neurons in the previous layer share the weights so that the same number of weights as the neurons in the next layer is used when connecting to the next layer.

The authors test the proposed model architecture's anomaly detection performance on the 11 types of short circuit faults identified [36], where an overall classification accuracy of 99.81% is achieved. However, they go on to state that the classification accuracy is not detailed enough to provide a conclusion as to whether the system is robust. They also go on to test the effect of current transformer saturation on the classification ability of the network, as well as voltage and current inversions. The results achieved for the stated cases are 99.4% and 97% respectively. The proposed network was also validated on real-world data, with an accuracy of 92% attained.

This paper [36] provides an objective view on the different methods used currently in power system fault detection and goes into depth on the prevalence of some time series image representation techniques over others but fails to provide an evidence-backed explanation as to why raw data is unsuitable for this use case only stating that "it is difficult to directly consider" [36] them for the fault detection and classification task. This directly contradicts the statement made about the difficulty of applying signal processing tasks due to the expert knowledge required. Moreover, not every test case was explored with this approach, which was identified by the authors [36] but this is to be expected since it is difficult to cover all fault types for such a complex system. On the other hand, a methodical approach was used to synthesise the proposed approach, and an evidence-backed conclusion was made to the effectiveness of their approach due to the various test cases applied and the comparison between traditional neural network models without the capsule integration. Furthermore, the weight shared capsule approach showed promise with its strong generalisation performance with the real-world dataset.

### D. LSTM and Capsule hybridisation

Recently, some works have been published combining the LSTM and Capsule networks to improve performance over current state-of-the-art techniques[39], [40]. One such work [41] proposes a rotating machinery fault diagnosis methodology utilising a CNN for feature extraction, a Bi-directional LSTM for denoising by dimensionality reduction and Capsules for their superior feature learning ability. The authors demonstrate the effectiveness of each proposed addition through a comparison of similar architectures with different NN combinations. The model diagnoses bearing faults in a supervised manner; this was demonstrated with an experiment on the Case Western Reserve University (CSWU) Bearing dataset [42], where raw vibration waveforms were used as the input data, and prelabelled classes as the output. The proposed model outperforms the current state-of-the-art with 98.95% accuracy whilst dramatically reducing training sample size from 5600, the sample size used by other compared methods, to just 150.

Another proposed method [43] combines the LSTM and Capsule layers in a single model for EEG emotion recognition. The authors propose a channel-wise attention mechanism using a CNN to prioritise the relevant EEG channels, Capsules to extract the spatial features and LSTM layers to extract the temporal features of the data. The model was tested on a public EEG dataset, where the state-of-the-art was considerably outperformed. It was noted that NN models that use Capsules consistently outperformed models using just CNNs in all three classification categories of valence, arousal and dominance. However, the authors noted the higher computational time involved with training Capsules as opposed to CNNs.

These examples in literature demonstrate the potential advantages of hybridising the LSTM and Capsule Networks, due to the advantages that the LSTM has with temporal feature learning and the Capsules spatial feature learning ability. Furthermore, the combination of the two layers has proven to be effective in some time series applications and has potential to be used in other architectures for different use cases.

To conclude, soft sensing methods of anomaly detection are more efficient and effective methods than hardware redundancy. However, with traditional methods it is difficult to accurately model systems without in depth knowledge of their dynamics and parameters, creating a barrier to flexible and accurate system modelling which is the basis of many anomaly detection systems. However, NNs provide a solution for this issue, providing a method of easily modelling system behaviour based of previously encountered data. Using NNs such as LSTM NNs for time series data learning, researchers have been able to accurately account for long term dependency in temporal data and create robust anomaly detection systems. However, this creates another issue with requiring access to large amounts of labelled data which is expensive and time consuming to produce. To avoid labelling data, some literatures have proposed unsupervised learning techniques such as the autoencoder which is able to learn data features by transforming it into a latent space representation. However, a comprehensive training set which is fully representative of the operation of the system or device being analysed is still required to utilise this technique, and generalisation performance is weak in many these methods. The Capsule was proposed to address issues layer to decide which values to update and a tanh layer creates a vector of candidate values to add to the cell state. The old state is then multiplied by $f_t$, the forget value and added to the candidate values scaled by the sigmoid. The "output" gate ,(5),(6), determines which part of the cell state to output using a sigmoid layer and is then multiplied by the cell state with a tanh layer applied to constrain the values between 1 and -1. The following equations formally define each gate:

TABLE I
LSTM EQUATIONS [4]

| Gate | Equation | |
|---|---|---|
| Forget | $f_t = \sigma(W_f \cdot [h_{t-1}, x_t] + b_f)$ | (1) |
| Replace | $i_t = \sigma(W_i \cdot [h_{t-1}, x_t] + b_i)$ | (2) |
| | $\tilde{C}_t = \tanh(W_C \cdot [h_{t-1}, x_t] + b_C)$ | (3) |
| | $C_t = f_t * C_{t-1} + i_t * \tilde{C}_t$ | (4) |
| Output | $o_t = \sigma(W_o \cdot [h_{t-1}, x_t] + b_o)$ | (5) |
| | $h_t = o_t * \tanh(C_t)$ | (6) |
| Symbol Definitions | $\sigma$=sigmoid function, $W_g$ = Weight of respective gate(g) neurons, $b_g$= Bias of respective gate(g) [f = forget gate, I = input gate, C = candidate gate, O = output gate] x=input at current timestep, h=output of previous LSTM cell | |

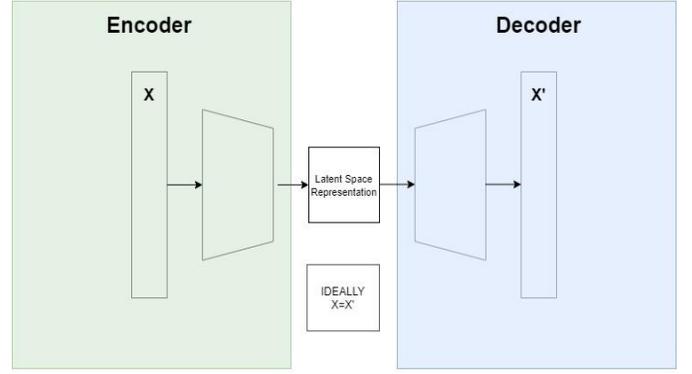

**Fig. 2.** Autoencoder architecture visualised [44]

### B. Capsule

The Capsule network (CapsNet) [14], is a novel neural network architecture designed to address the issues the convolutional neural network has with spatial context. It does this by 'encapsulating' the spatial information between the variables using vectors which allows the neural network to learn the distances between the identified features as well as the classification of the features.

A capsule differs from the traditional artificial neuron in various ways: A traditional neuron receives scalar inputs; performs the weighted sum of the aforementioned scalars; applies an activation function and outputs a scalar dependant on the weights and biases that it has adopted through training. A capsule on the other hand, whilst operating in a similar fashion, slightly differs from the internal operation and the representation of the values that it receives. A capsule receives a vector input, where the input denotes the probability of occurrence as well as orientation and other spatial features not captured by a scalar value. It applies an "affine transformation" which is essentially a transformation matrix weight that replaces the traditional scalar weight; this operation is formally defined in (7)[14]:

$$\hat{u}_{j|i} = W_{ij} u_i.$$ (7)

This transformation matrix is used to represent the spatial context that is missing from the traditional method of weight application. The weighted sum of these vectors is calculated using (8) [14]:

$$s_j = \sum_i c_{ij} \hat{u}_{j|i}.$$ (8)

In order to preserve the vector information that is input to the capsule, a new type of activation is proposed known as the non-linear "squashing" function. This operates similar to the normal

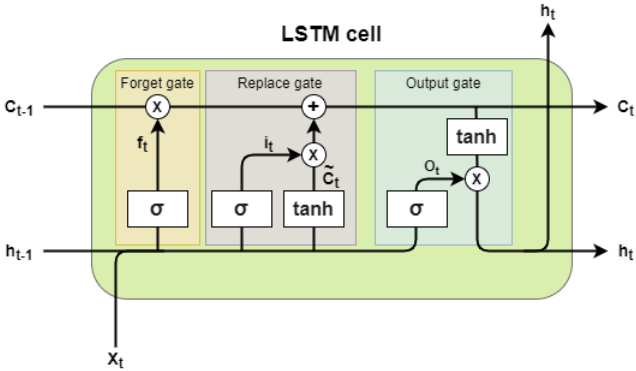

**Fig. 1.** Visualisation of LSTM cell [4], where: $\sigma$=sigmoid layer, tanh= tanh layer, $f_t$=forget gate, $i_t$=input gate, $\tilde{C}_t$=candidate gate, $O_t$=output gate

with training efficiency and the shortcomings of traditional NNs with learning spatial context of data. This paper further explores these qualities found in Capsules by hybridising them with LSTMs in a NN, and addresses issues found with learning multivariate data with single channel NNs.

### III. PROPOSED NEURAL NETWORK MODEL

#### A. LSTM Cell

The LSTM network, proposed initially in 1997 by Hochreiter and Schmidhuber [4] but popularised recently by its widespread usage in commercial environments, is a popular iteration of the RNN that can overcome the vanishing gradient issue and allows for the learning of long-term dependency. It does this using a specialised architecture that integrates "gates" to the architecture to allow the cell state to forget values and replace values, then decide which values to output and send to the next cell. A visualisation of this architecture can be seen in Fig. 1, and the equations shown in TABLE I.

The architecture can be described as follows: The "forget" gate, (1), uses a sigmoid layer to determine which irrelevant data in the cell state to remove, where a value of 1 would keep all of the data and a 0 would completely erase the information. The "replace" gate ,(2),(3),(4), is used to decide which values in the cell state to update. This gate operates by using a sigmoid activation functions discussed previously by squashing the output between 0 and 1 but does so in a way that is able to preserve the length and spatial information of the input values, so that a long vector will shrink to a value just below 1 and shorter vectors are shrunk to near 0 [14]. Equation (9) [14] formally defines this operation:

$$v_j = \frac{||s_j||^2}{1+||s_j||^2} \cdot \frac{s_j}{||s_j||}.$$ (9)

### C. Autoencoder Structure

An Autoencoder is a variant of neural network architecture that aims to learn a compressed representation of the input data and copy it to the output. A compressed representation is used so that the model does not learn the noise in a data representation but only the main shapes and features of the data. An autoencoder is composed of 2 sections: An encoder and a decoder. The encoder part is used to transform the input data into a latent space representation through dimensionality reduction, which the decoder part then learns and decodes back into the input data with reduced accuracy and hence noise. A visualization of this can be seen in Fig. 2.

A formal definition of the Autoencoder operation is provided in (10) and (11) [44]:

$$f(x) = latent\ space$$ (10)

$$g(f(x)) = x'$$ (11)

The idea is to reduce the layer width of the middle layers so that the neural network compresses the input instead of just learning the exact representation: this is known as an undercomplete Autoencoder. However, learning data that is too compressed would reduce the accuracy of the reconstruction, so when training the network, the aim is to balance the denoising ability with the accuracy of reconstruction. This is determined by the reconstruction loss, and the aim of training this type of network is to minimise this loss whilst maintaining a good generalisation performance.

This type of neural network is typically used for unsupervised deep learning, as the inputs are being copied to the outputs with no labelling required, which is useful for the use case that this paper explores.

### D. Proposed Layer Architecture

The proposed layer architecture is illustrated in Fig. 3 As shown in the figure, the number of input branches is entirely dependent on the number of features present in the data.

Each individual feature is first encoded through an LSTM layer for dimensionality reduction and is output as a 2D vector. This dimensionality reduction is carried out to reduce the number of degrees of freedom in the model so that the risk of overfitting on data is reduced. Additionally, representing the data in latent space also helps the NN to learn data features more easily. The Repeat Vector layer transforms the 2D tensor input back to a 3D tensor by repeating the fixed length vector n number of times; in this case the number of repetitions is set to be equal to the number of timesteps being input in the system. The 3D tensor is then input into a Capsule layer, where the number of Capsules in equal to the number of timesteps, and the width of the Capsule is equal to the LSTM hidden layer width. The output of each branch is then concatenated into a single vector which is input into a Capsule layer with an equal number of Capsules as the previous Capsule layers but with a width equal to the product of the width of the previous Capsules and the number of branches. The Time Distributed layer is then used to apply a Dense layer to each vector in the 3D output. Since this is an Autoencoder, the output should be equal to the input vectors merged into one vector with all the input features present.

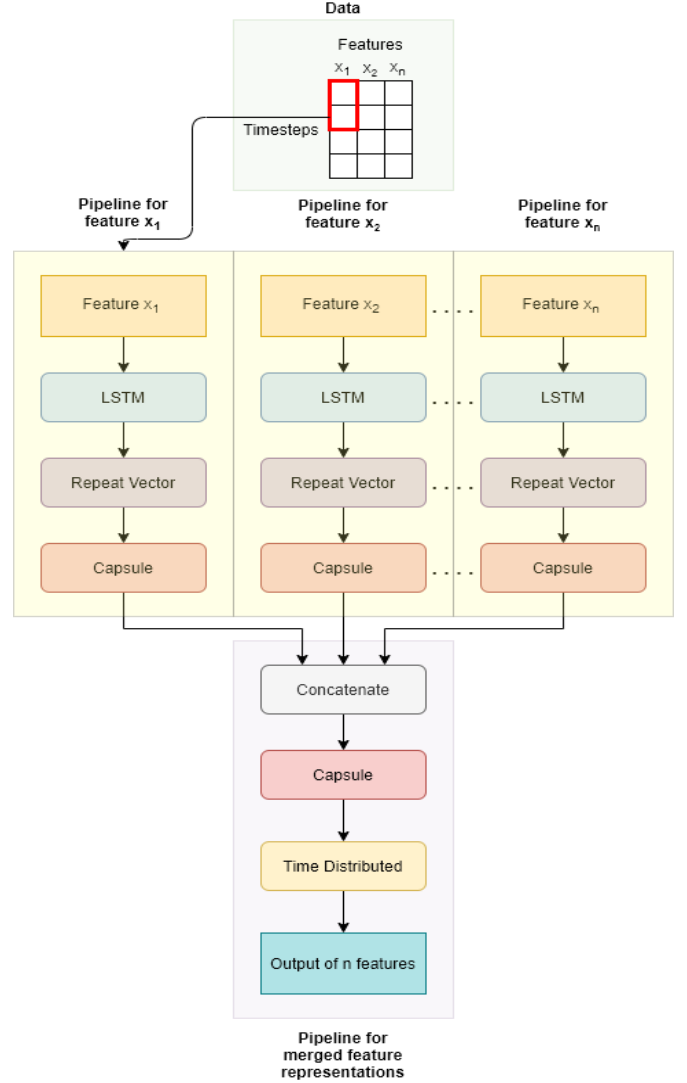

**Fig. 3.** Architecture proposed by this paper

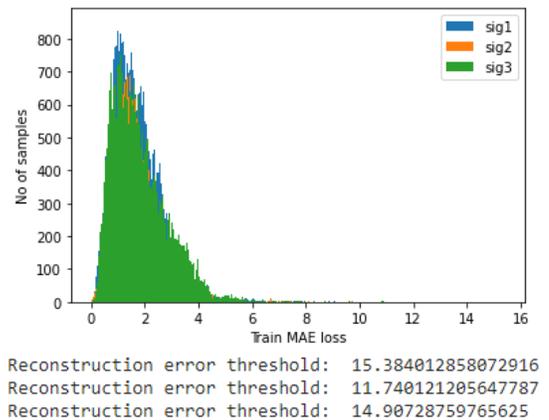

**Fig. 4.** An example of a frequency plot of the MAEs for the drone dataset (where the signal sources are unknown), and the corresponding error thresholds for feature 1, feature 2 and feature 3 in Fig. 9.

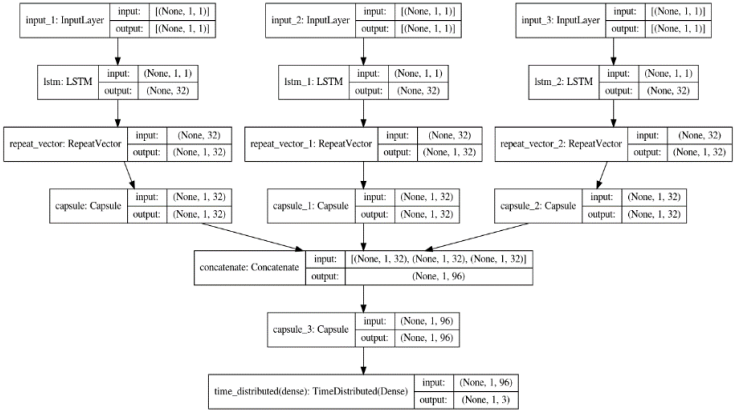

**Fig. 5.** Design A (The proposed design) – Branched Inputs, LSTMCaps Autoencoder Network

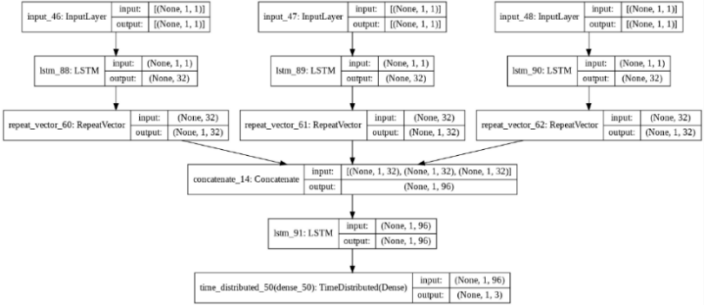

**Fig. 6.** Design B – Branched Inputs, LSTM Autoencoder Network with no Capsules

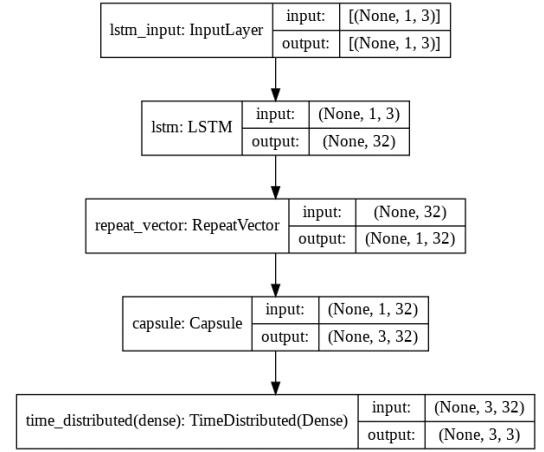

**Fig. 7.** Design C – Single Input, LSTMCaps Autoencoder Network

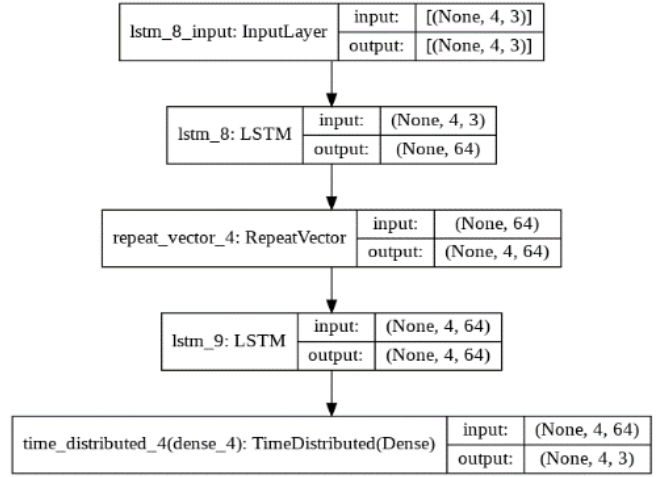

**Fig. 8.** Design D – Single Input, LSTM Autoencoder Network

### E. Fault Detection method

To be able to detect a fault, the neural network will be trained to reconstruct healthy data. This can be done when fitting by setting the input and the output as the same dataset, which would be the healthy operation of the data. While technically this is an unsupervised task as the initial data provided is unlabeled so the condition of the data is unknown, it can be framed as a supervised learning task using this method.

After this, the reconstruction error can be found using the Mean Absolute Error (MAE) of the training predictions. The maximum prediction error for the training set can be used as the reconstruction error threshold, which essentially means that the worst prediction case is being used as the threshold initially so that when applying the system to more data from the system being analysed, any predictions outside this value will be more likely to be an anomaly. The sensitivity of the anomaly detection can be adjusted by changing the threshold value, so this will be experimented with in order to find the optimal value that will minimise the false positive and false negative rate. Furthermore, each data feature will have its own error threshold to maximise the accuracy of detection as the neural network may perform better on some features than others. An example of a threshold calculation can be seen in Fig. 4. The main aim of the training process is to minimise the loss and standard deviation of this plot so that the system is able to make more confident and sensitive anomaly predictions.

### IV. EXPERIMENTATION

To demonstrate the effectiveness of the contributions of this paper, four different NN models have been compared:
- For the first experiment, the effectiveness of CapsNet integration is explored in terms of training performance

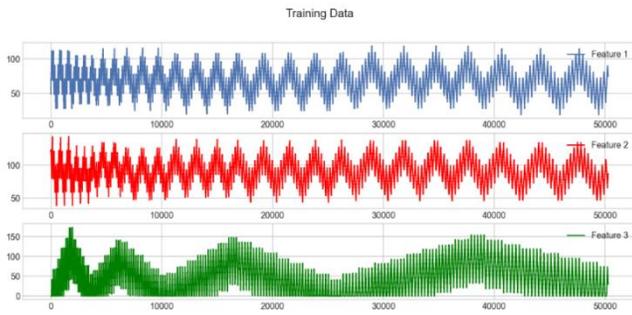

**Fig. 9.** Training data subset from the reference drone with the 3 outputs shown

and MSE of prediction on test data. The proposed model (Design A, Fig. 5) is compared to a similar model with the Capsule layers removed and replaced with LSTM layers (Design B, Fig. 6).

- For the second experiment, the effectiveness of the branched model was observed. This was demonstrated by synthesising a non-branched model with the same layer structure as the branched model (Design C, Fig. 7).

- The final experiment will demonstrate the effectiveness of both additions being used simultaneously. This will be shown by using a non-branched model consisting of just LSTM layers (Design D, Fig. 8).

For the following experiments, each model was adjusted so that the trainable parameters are similar to the LSTMCaps model for experimental consistency. The models were then trained on the datasets with equal training iterations (epochs) 5 times each, and an average was taken. This was to observe the training stability of the respective models.

The final training and validation loss values will be used to measure the training efficiency of each NN model tested, and the Mean Squared Error (MSE), Mean Absolute Error (MAE) and F1 score of testing data predictions will be used to measure the prediction accuracy of the models respectively.

*A. Datasets*

*Drone Data*

The drone dataset was acquired from previous work conducted by a researcher in the same department. The dataset consists of 3 output features from unknown sensors. 3 subsets of data were provided, the sample sizes and respective visualisations are outlined below:

*Reference device:*

- Sample size: 600 secs (Fig. 9)

- Sample size: 30 secs (Fig. 10)

*Test Device:*

- Sample size: 30 secs (Fig. 11)

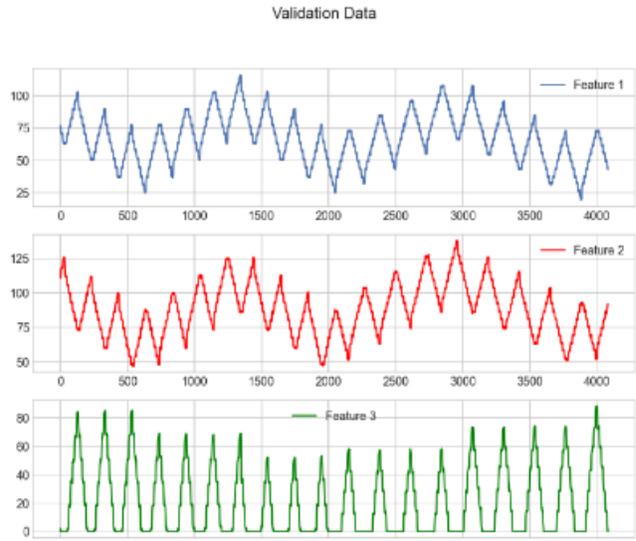

**Fig. 10.** Validation data subset from the reference drone with the 3 outputs shown

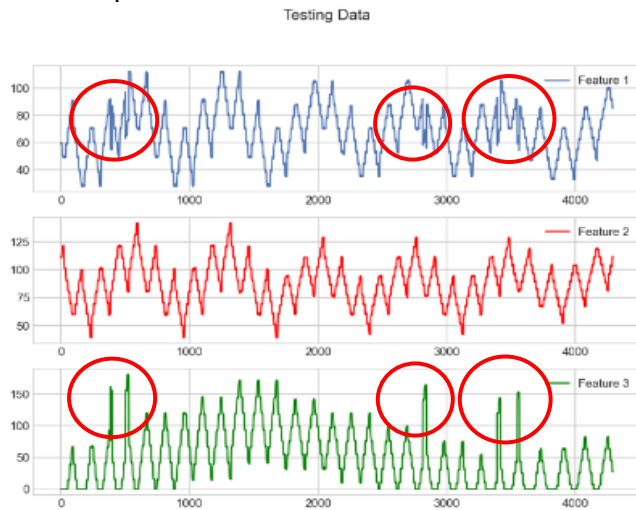

**Fig. 11.** Testing data subset from the test drone with the 3 outputs shown, and anomalies circled in red

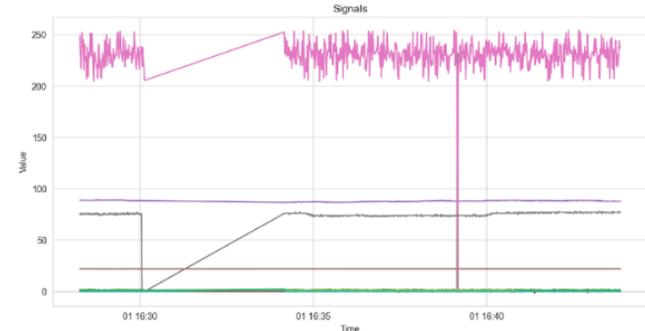

**Fig. 12.** Plot of a subset of data from the SKAB anomaly benchmark

TABLE III
RESULTS FOR TRAINING FOR EACH NEURAL NETWORK MODEL USING HYPERPARAMETERS FROM TABLE II

| Model | Design A: Branched LSTMCaps | Design B: Branched LSTM | Design C: Non-Branched LSTMCaps | Design D: Non-Branched LSTM |
|---|---|---|---|---|
| Training Plot | 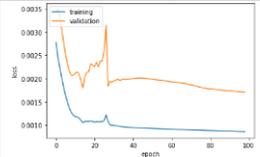 | 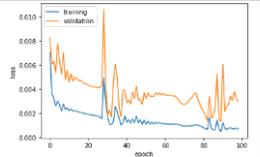 | 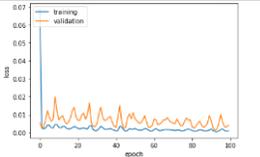 | 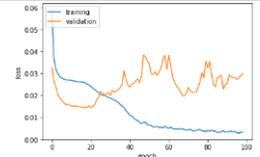 |
| Avg Final Train Loss | 0.0013 | 0.0012 | 0.0019 | 0.0052 |
| Avg Final Val loss | 0.0017 | 0.0030 | 0.0041 | 0.0299 |
| % Overfitting | 31 | 150 | 116 | 475 |
| % Val loss improvement from non-Caps | 43 | N/A | 86 | N/A |

TABLE II
INITIAL HYPERPARAMETERS USED TO TRAIN MODELS

| Hyperparameter | Value | Hyperparameter | Value |
|---|---|---|---|
| Epochs | 100 | Loss Function | MSE |
| Optimiser | Adam | Dropout rate | 0.2 |
| Learning rate | 0.001 | Batch size | 64 |
| Time Steps | 64 | Branched layer width | 32 |
| Capsule Activation | squash | LSTM Activation | tanh |

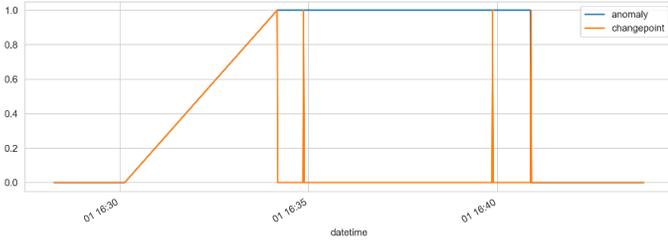

**Fig. 13.** Binary plot of the respective anomalies and changepoints in Fig. 12

As shown in Fig. 11, it is clear to see that there are anomalies from the test device in feature 1 and feature 3 at the same points temporally. The reason for the anomalous data stems from malicious code affecting the drone controls causing the direction of the drone to differ from the intended direction input by the drone operator. Since the data was received unlabelled, the anomalies were manually labelled so that a measure of the anomaly detection performance of each NN model could be attained. The metric used for this is the F1 score, which is defined in (12):

$$F_1 = \frac{TP}{TP + \frac{1}{2}(FP + FN)} \ , \qquad (12)$$

where $F_1$ = F1 Score, TP = True Positive, FP = False Positive, FN = False Negative.

*SKAB Anomaly Benchmark*

The SKAB anomaly detection benchmark [45] is a public benchmark available online used for offline outlier detection and changepoint detection testing. The benchmark consists of 35 subsets of data from a water circulation system which contain 8 features each from different sensors in the system. The test is conducted by looping through each subset, training the neural network on a slice of clean data from the subset then testing it on labelled anomalies that were simulated with the test rig. The metrics used to gauge the effectiveness of the anomaly detection are the F1 score (12) and the NAB Changepoint

Metric [46]. Fig. 12 illustrates a subset of data from the benchmark, and Fig. 13 shows the plot for the anomalies in the data.

*B. Data Pre-processing*

Data pre-processing can be segmented into 4 sections: Cleaning, Integration, reduction, and transformation. For the datasets acquired, most of the pre-processing procedure was not essential as the data was acquired in a format that implied that the cleaning and integration had already been carried out. Moreover, the data did not require reduction since time series data is sequential so removing or shuffling the dataset would compromise the integrity of the readings. However, it was necessary for the data to be transformed; this is an integral part of data pre-processing and is carried out due to the benefits it can have on the performance on the neural network with the speed of convergence when training as well as performance. To improve the neural network performance, it is common practice to rescale the data. For these datasets, the type of rescaling chosen was Z-Score Normalisation (13):

$$z = \frac{(X - \mu)}{\sigma}, \qquad (13)$$

where X = un-normalised data point, μ = mean of the dataset, σ = standard deviation of the dataset and z = normalised data point. This operation normalises the dataset so that the mean is equal to 0 and the standard deviation is equal to 1.

*C. Experiment 1: Drone Data*

This experiment aims to explore the training capability of the proposed NN architecture (Design A, Fig. 5) by making a

TABLE IV

BEST TEST RESULTS FOR ANOMALY DETECTION FROM 5 RUNS USING OPTIMISED HYPERPARAMETERS FOR EACH NN DESIGN

| Model | Trainable Parameters | MSE | MAE Threshold | | | Std Dvn of thresholds | Precision | Recall | F1 |
|---|---|---|---|---|---|---|---|---|---|
| | | | Feature 1 | Feature 2 | Feature 3 | | | | |
| *Design A: Branched LSTMCaps* | *25,635* | *0.0244* | *0.8126* | *0.9764* | *1.1383* | *0.1330* | *0.8819* | *0.6633* | *0.7465* |
| Design B: Branched LSTM | 24,243 | 0.0145 | 0.6222 | 0.9034 | 0.9389 | 0.1417 | 0.8452 | 0.3333 | 0.3949 |
| Design C: Non-Branched LSTMCaps | 26,183 | 0.0313 | 0.5367 | 1.0195 | 0.8689 | 0.2017 | 0.6219 | 0.5533 | 0.5143 |
| Design D: Non-Branched LSTM | 25,338 | 0.0309 | 0.9435 | 0.7682 | 1.5487 | 0.3344 | 0.7224 | 0.5767 | 0.6406 |

TABLE V

AVERAGE TEST RESULTS FOR ANOMALY DETECTION FROM 5 RUNS USING OPTIMISED HYPERPARAMETERS FOR EACH NN DESIGN

| Model | Trainable Parameters | MSE | MAE Threshold | | | Std Dvn of thresholds | Precision | Recall | F1 |
|---|---|---|---|---|---|---|---|---|---|
| | | | Feature 1 | Feature 2 | Feature 3 | | | | |
| *Design A: Branched LSTMCaps* | *25,635* | *0.0187* | *0.5998* | *0.6842* | *0.8090* | *0.1119* | *0.8334* | *0.5666* | *0.6415* |
| Design B: Branched LSTM | 24,243 | 0.0118 | 0.8371 | 0.8588 | 0.8627 | 0.1081 | 0.8706 | 0.3387 | 0.3680 |
| Design C: Non-Branched LSTMCaps | 26,183 | 0.0587 | 1.4832 | 1.4752 | 1.5623 | 0.1161 | 0.7965 | 0.3253 | 0.3254 |
| Design D: Non-Branched LSTM | 25,338 | 0.0273 | 1.0303 | 0.6698 | 1.3450 | 0.2965 | 0.5141 | 0.4460 | 0.4663 |

comparison with the non-hybridised non-branched LSTM NN (Design D, Fig. 8), the non-branched hybridised LSTMCaps NN (Design C, Fig. 7) and the branched non-hybridised LSTM NN (Design B, Fig. 6). The models are first trained with non-optimised hyperparameters, then each optimised for the drone dataset and tested for their anomaly detection capabilities.

The NN models were first tested with the default recommended hyperparameters in TensorFlow documentation and literature; this was to purely observe the raw effect of the inclusion of the Capsule layer as well as the introduction of the branched architecture on the training performance. The hyperparameters used are shown in Table II. Each model was trained 5 times on the 5-minute subset from the reference device and the training and validation loss scores were recorded. An average was taken of these values for testing rigour: the results are shown in Table III, as well as the improvement in training performance with the inclusion of the Capsule Layer. The training plot is also illustrated so that the stability of the training can be better visualised. Additionally, the percentage of overfitting is shown, which refers to the percentage difference between the training and validation losses.

The neural network models were then optimised using an iterative testing method so that the effect of changing each hyperparameter value can be seen and hence from this, the most optimal configuration of the hyperparameters for each model can be found.

The values that were monitored were the training and validation loss, and the training time. The loss value was chosen as the metric to gauge the effectiveness of the system due to the nature of the outlier detection technique. Since the anomaly threshold is calculated using the prediction residual, a low loss value allows for a lower threshold for the loss when testing the model, which can potentially lead to a more sensitive and accurate fault detection system. The validation loss scores were considered with more weight than the training loss when quantitively analysing the system as they were used to determine the error threshold, as well as them being a better indicator of the generalisation ability of the NN.

The optimal hyperparameters were then implemented into the proposed models for further testing. Before conducting the testing, each NN model was adjusted so that all networks being trained have a similar number of parameters for the purpose of experimental rigour. This will reduce the difference between each model so that the effect of the proposed architecture and hybridisation can be better observed on anomaly detection performance.

Each model was trained 5 times, and for each individual training procedure the prediction MSE and MAE thresholds were recorded, as well as the standard deviations of the latter to observe the consistency of training for each feature. Using the thresholds, the NNs were made to predict the test data, and any predictions exceeding the thresholds set were outlined as anomalies. The predicted anomalies were then compared to the real anomalies labelled during data analysis, and the precision, recall and F1 scores were calculated for each NN. The best score attained by each NN model is depicted in Table IV, and the average score over the 5 runs in Table V.

The results in Table III clearly show an improvement in performance with the proposed additions. The addition of the Capsule layer to the non-branched model variant using non-optimised hyperparameters shows a clear improvement in the training and validation losses respectively. Training results using non-optimised hyperparameters have an overall more stable training procedure; evidence for this is shown in the training plots accompanying the results for Design A and Design C. With the addition of the branched inputs, there is a significant improvement in performance in both the hybridised and non-hybridised models with non-optimised hyperparameters. The branched model shows a clear reduction in overfitting from 475% to 150% without Capsule layers and from 115.79% to 30.77% with the Capsule layer without hyperparameter optimisation.

After optimising each NN model hyperparameters on the dataset, the results in Table IV and Table V show that the proposed model, Design A, performs better than the other models tested with anomaly detection with an average F1 score of 0.64, and a best F1 score of 0.75. However, the non-branched

TABLE VI

AVERAGE OUTLIER DETECTION SCORES FROM 5 TEST ITERATIONS FOR EACH ANOMALY DETECTION METHOD

| Algorithm | F1 | FAR, % | MAR, % |
|---|---|---|---|
| *Perfect score* | *1* | *0* | *0* |
| **LSTMCaps** | **0.74** | **21.66** | **18.74** |
| MSET[48] | 0.73 | 20.82 | 20.08 |
| *LSTMCapsV2* | *0.71* | *14.45* | *30.86* |
| MSCRED [49] | 0.7 | 16.82 | 31.28 |
| Conv-AE [50] | 0.66 | 5.57 | 46.16 |
| LSTM [51] | 0.65 | 14.89 | 39.4 |
| LSTM-AE [52] | 0.64 | 14.81 | 39.5 |
| LSTM-VAE [53] | 0.56 | 9.04 | 54.75 |
| Autoencoder [54] | 0.45 | 7.52 | 66.59 |
| Isolation forest [47] | 0.4 | 6.86 | 72.09 |
| *Null score* | *0* | *100* | *100* |

TABLE VII

AVERAGE CHANGEPOINT DETECTION SCORES FROM 5 TEST ITERATIONS FOR EACH ANOMALY DETECTION METHOD

| Algorithm | NAB (standard) | NAB (lowFP) | NAB (LowFN) |
|---|---|---|---|
| *Perfect score* | *100* | *100* | *100* |
| Isolation forest [47] | 37.53 | 17.09 | 45.02 |
| *LSTMCapsV2* | *27.39* | *17.08* | *31.13* |
| LSTM | 26.61 | 11.78 | 32 |
| MSCRED [49] | 26.13 | 17.81 | 29.53 |
| LSTM-AE [52] | 22.97 | 20.95 | 23.93 |
| **LSTMCaps** | **21.58** | **5.12** | **27.49** |
| LSTM-VAE [53] | 21.09 | 17.52 | 22.73 |
| Autoencoder [54] | 15.65 | 0.48 | 21 |
| MSET[48] | 12.71 | 11.04 | 13.6 |
| Conv-AE [50] | 11.12 | 10.35 | 11.77 |
| *Null score* | *0* | *0* | *0* |

TABLE VIII

BEST OUTLIER DETECTION SCORES OUT OF 5 TEST ITERATIONS FOR EACH ANOMALY DETECTION METHOD

| Algorithm | F1 | FAR, % | MAR, % |
|---|---|---|---|
| *Perfect score* | *1* | *0* | *0* |
| **LSTMCaps** | **0.74** | **21.5** | **18.74** |
| MSET [48] | 0.73 | 20.82 | 20.08 |
| *LSTMCapsV2* | *0.71* | *14.51* | *30.59* |
| MSCRED [49] | 0.7 | 16.2 | 30.87 |
| LSTM [51] | 0.67 | 15.42 | 36.02 |
| Conv-AE [50] | 0.66 | 5.58 | 46.05 |
| LSTM-AE [52] | 0.65 | 14.59 | 39.42 |
| LSTM-VAE [53] | 0.56 | 9.2 | 54.81 |
| Autoencoder [54] | 0.45 | 7.55 | 66.57 |
| Isolation forest [47] | 0.4 | 6.86 | 72.09 |
| *Null score* | *0* | *100* | *100* |

TABLE IX

BEST CHANGEPOINT DETECTION SCORES OUT OF 5 TEST ITERATIONS FOR EACH ANOMALY DETECTION METHOD

| Algorithm | NAB (standard) | NAB (lowFP) | NAB (LowFN) |
|---|---|---|---|
| *Perfect score* | *100* | *100* | *100* |
| Isolation forest [47] | 37.53 | 17.09 | 45.02 |
| *LSTMCapsV2* | *27.77* | *17.14* | *31.59* |
| LSTM [51] | 26.76 | 12.92 | 31.93 |
| MSCRED [49] | 24.99 | 17.9 | 27.94 |
| LSTM-AE [52] | 24.77 | 22.69 | 25.75 |
| **LSTMCaps** | **24.02** | **8.14** | **29.60** |
| LSTM-VAE [53] | 21.92 | 18.45 | 23.59 |
| Autoencoder [54] | 16.27 | 1.04 | 21.62 |
| MSET [48] | 12.71 | 11.04 | 13.6 |
| Conv-AE [50] | 11.21 | 10.45 | 11.83 |
| *Null score* | *0* | *0* | *0* |

TABLE X

SCALED AVERAGE OF AVERAGE F1 AND NAB SCORES FROM TABLE VI AND TABLE VII RESPECTIVELY

| Algorithm | Scaled Average of Average Score |
|---|---|
| *Perfect score* | *1* |
| *LSTMCapsV2* | *0.49195* |
| MSCRED [49] | 0.48065 |
| **LSTMCaps** | **0.4779** |
| LSTM [51] | 0.45805 |
| LSTM-AE [52] | 0.43485 |
| MSET [48] | 0.42855 |
| Isolation forest [47] | 0.38765 |
| Conv-AE [50] | 0.3856 |
| LSTM-VAE [53] | 0.38545 |
| Autoencoder [54] | 0.30325 |
| *Null score* | *`0* |

TABLE XI

SCALED AVERAGE FROM BEST F1 AND NAB SCORES FROM TABLE VIII AND TABLE IX RESPECTIVELY

| Algorithm | Scaled Average of Best Score |
|---|---|
| *Perfect score* | *1* |
| *LSTMCapsV2* | *0.49385* |
| **LSTMCaps** | **0.4901** |
| MSCRED [49] | 0.47495 |
| LSTM [51] | 0.4688 |
| LSTM-AE [52] | 0.44885 |
| MSET [48] | 0.42855 |
| LSTM-VAE [53] | 0.3896 |
| Isolation forest [47] | 0.38765 |
| Conv-AE [50] | 0.38605 |
| Autoencoder [54] | 0.30635 |
| *Null score* | *0* |

LSTM model, Design D, performs better than both the non-branched hybridised and the branched non-hybridised models. This was found to be the case due to the technique used for anomaly detection: With a higher MSE, Design D (the standard LSTM AE, Fig. 8) did not learn the data features as accurately as Design B (the branched LSTM variant, Fig. 6), which in this case was more beneficial for anomaly detection since data is more likely to be flagged as an outlier. Whilst this resulted in a higher F1 score, the precision of the model is weaker in comparison to both branched variants. In the case of Design B, the average MSE of prediction was the lowest out of all the models, which did not work to its favour during anomaly detection with recall but resulted in a higher precision.

By utilising the hybridisation in the branched input model, the best performance was achieved across all the metrics tested, with both optimised and non-optimised hyperparameters. Furthermore, minimal overfitting was observed when training with unoptimized hyperparameters. Consequently, it can be said that minimal hyperparameter optimisation is required when using this model architecture as these results show a resilience to overfitting and relatively strong performance when applying the network on multivariate data without hyperparameter tuning.

The results attained show that both the hybridisation of the Capsule and LSTM layers and the branched input model structure are both effective methods of improving the performance of the neural network with multivariate data, especially when used in conjunction with each other. To further substantiate this performance, the proposed model was tested against common state-of-the-art anomaly detection methods on an open-source benchmark.

### D. Experiment 2: SKAB Anomaly benchmark

This experiment aims to compare the anomaly detection and changepoint detection performance of state-of-the-art unsupervised anomaly detection methods with the proposed NN model. A selection of NNs and ML based fault detection methods were chosen to compare on the benchmark with minimal hyperparameter optimisation applied.

The same testing procedure utilised in the SKAB benchmark's GitHub repo [45] was used to test the proposed model architecture. The model was trained with 100 epochs on a subset from each dataset with early stopping set at a patience of 20, and then tested on the remainder of the dataset. The F1 scores and NAB scores achieved for each dataset are averaged, which gives the final score of the benchmark. Each model compared was also tested on the same computer for experimental rigour. The results in Table VI and Table VII depict the average outlier detection score and the changepoint detection scores over 5 test iterations respectively, and the results in Table VIII and Table IX show the best NN performance in a single test over the outlier and changepoint scores respectively.

To better conclude the effectiveness of each anomaly detection method over both the F1 and NAB scores simultaneously, a scaled average of both metrics was calculated. This was done by scaling the NAB score between 0 and 1 and averaging the F1 score and scaled NAB scores. The results in Table 1 and Table 2 show the scaled average of the F1 and NAB score of the average performance and best performance respectively.

While testing it was found that there was an inversely proportional relationship with outlier detection score and changepoint detection score. This meant that hyperparameters optimal for a good F1 score would not perform as well in the NAB score. To demonstrate this, the hyperparameters of the LSTMCaps NN were slightly adjusted to achieve a better score in the changepoint detection benchmark, at the expense of a slightly lower outlier detection score. This NN configuration is labelled as **LSTMCaps V2** in the results shown in Table VI to Table XI.

The results in Table VI and Table VIII show that as an outlier detector, the proposed LSTMCaps NN achieves the best F1 score and the lowest False Negative rate out of the models tested. It also achieves the second highest False Positive rate out of the models. In terms of changepoint detection, the results in Table VII and Table IX indicate that the original configuration does not perform as well, coming 5th out of the 9 methods tested. However, with a slight adjustment to the hyperparameters, the LSTMCapsV2 NN was able to come 2nd out of the 9 methods tested in both the outlier detection and changepoint detection scores and performs better than all other NN based methods in the latter. Similar outcomes can be seen for the best performing test iteration, with no improvement in relation to the other NNs and ML methods. The scaled average results in Table X and Table XI show that the LSTMCapsV2 configuration is overall the best performing method over the two metrics tested.

From this test, it can be concluded that for single datapoint outlier detection, the proposed LSTMCaps branched architecture provides state-of-the-art performance. However, while the changepoint detection performance is superior to other NNs with the right adjustments to the hyperparameters, significantly better performance can be attained from non-NN based algorithms, such as the Isolation Forest algorithm [47].

## V. DISCUSSION

Across the experiments conducted, it is clear to see that both the inclusion of the Capsule Network and the branched input architecture is integral to the improvement of the performance of the Capsule Network in terms of training and anomaly detection. The evidence for this is shown clearly across the experiments, where with standard LSTM AEs, the training and anomaly detection performance is significantly weaker than with the proposed NN.

The experimental results further suggest that the Capsule Network is most effective in the training phase. Generally, it was found that models which included Capsules were training more efficiently, reaching the local minima at a faster rate in relation to networks without Capsules. Most importantly, the results in Table III for training using non-optimised hyperparameters suggest that with the use of Capsules, the hyperparameter optimisation procedure can be simplified considerably due to the lack of overfitting during training on the NN models with Capsules integrated.

One significant strength of the proposed LSTMCaps NN is its ability to learn separate data features effectively in comparison to a standard single channel NN. This is shown by the difference in standard deviation in the MAE thresholds in Table IV when conducting the anomaly detection test on the drone data. This is further substantiated with the anomaly detection performance on the SKAB anomaly benchmark, which contains a larger number of more complex features than the drone data. Here it is clear to see the advantage that having separate input branches per feature brings.

## VI. CONCLUSION AND FUTURE WORKS

This paper proposed a novel hybridisation of the LSTM and Capsule Networks in a branched architecture to address the issues found in the literature review with training performance of NNs, specifically on multivariate data. The motivation for this research stemmed from the growing demand for more effective unsupervised data analysis techniques regarding outlier and anomaly detection for use in industrial and commercial environments with large datasets to assist in the advancement of Industry 4.0, the automation of industrial processes.

The proposed NN was tested first in its training performance with no hyperparameter optimisation and compared to non-hybridised and non-branched variants of the NN, where it was found that the proposed NN can train more efficiently over a smaller number of epochs in comparison to the variants with no capsules integrated in the NN, and significantly reduces overfitting. After conducting hyperparameter optimisation, the NNs were retested, this time for their anomaly performance ability using an unsupervised method of reconstructing the data and using the MAE any data outlying from the expected shape in the training data. The results of this test concluded that the proposed NN performs better than the other variants tested as a result of the proposed additions and changes to the NN architecture. To substantiate these results, the proposed NN was tested against other state-of-the-art anomaly detection methods on the SKAB anomaly detection benchmark, where with slight hyperparameter adjustments the proposed method was able to perform better than all other methods tested for outlier detection and performed better than all other NN based methods in changepoint detection, only being outperformed by the Isolation Forest algorithm in the latter.

Whilst the proposed NN operated exclusively on raw data, it was found in the literature review that with different representations of data, the prominence of data features can be increased which in turn can help to improve the performance of unsupervised anomaly detection. Furthermore, the use-cases for the proposed NN were not fully explored, so future works will be exploring the use of different data representations and different unsupervised anomaly detection methods, including the grouping of encountered anomalies in an unsupervised manner.

## REFERENCES

[1]  J. Lee, H. A. Kao, and S. Yang, "Service innovation and smart analytics for Industry 4.0 and big data environment," *Procedia CIRP*, vol. 16, pp. 3–8, 2014, doi: 10.1016/j.procir.2014.02.001.

[2]  E. Dubrova, "Hardware Redundancy," in *Fault-Tolerant Design*, New York, NY: Springer New York, 2013, pp. 55–86. doi: 10.1007/978-1-4614-2113-9_4.

[3]  M. Paliwal and U. A. Kumar, "Neural networks and statistical techniques: A review of applications," *Expert Systems with Applications*, vol. 36, no. 1, pp. 2–17, 2009, doi: 10.1016/j.eswa.2007.10.005.

[4]  S. Hochreiter and J. Schmidhuber, "Long Short-Term Memory," *Neural Computation*, vol. 9, no. 8, pp. 1735–1780, 1997, doi: 10.1162/neco.1997.9.8.1735.

[5]  A. Nanduri and L. Sherry, "Anomaly detection in aircraft data using Recurrent Neural Networks (RNN)," *ICNS 2016: Securing an Integrated CNS System to Meet Future Challenges*, pp. 1–8, 2016, doi: 10.1109/ICNSURV.2016.7486356.

[6]  T. Ergen and S. S. Kozat, "Unsupervised anomaly detection with LSTM neural networks," *IEEE Transactions on Neural Networks and Learning Systems*, vol. 31, no. 8, 2020, doi: 10.1109/TNNLS.2019.2935975.

[7]  M. Zhang, W. Li, and Q. Du, "Diverse region-based CNN for hyperspectral image classification," *IEEE Transactions on Image Processing*, vol. 27, no. 6, pp. 2623–2634, 2018, doi: 10.1109/TIP.2018.2809606.

[8]  Q. Li, W. Cai, X. Wang, Y. Zhou, D. D. Feng, and M. Chen, "Medical image classification with convolutional neural network," *2014 13th International Conference on Control Automation Robotics and Vision, ICARCV 2014*, vol. 2014, no. December, pp. 844–848, 2014, doi: 10.1109/ICARCV.2014.7064414.

[9]  A. Dingli and K. S. Fournier, "Financial time series forecasting - a deep learning approach," *International Journal of Machine Learning and Computing*, vol. 7, no. 5, pp. 118–122, 2017, doi: 10.18178/ijmlc.2017.7.5.632.

[10]  C. Y. Hsu and W. C. Liu, "Multiple time-series convolutional neural network for fault detection and diagnosis and empirical study in semiconductor manufacturing," *Journal of Intelligent Manufacturing*, vol. 32, no. 3, pp. 823–836, 2021, doi: 10.1007/s10845-020-01591-0.

[11]  S. Mukhopadhyay and M. Litoiu, "Fault Detection in Sensors Using Single and Multi-Channel Weighted Convolutional Neural Networks," pp. 0–7, 2020.

[12]  M. Canizo, I. Triguero, A. Conde, and E. Onieva, "Multi-head CNN–RNN for multi-time series anomaly detection: An industrial case study," *Neurocomputing*, vol. 363, pp. 246–260, Oct. 2019, doi: 10.1016/j.neucom.2019.07.034.

[13]  T. Y. Kim and S. B. Cho, "Web traffic anomaly detection using C-LSTM neural networks," *Expert Systems with Applications*, vol. 106, pp. 66–76, Sep. 2018, doi: 10.1016/J.ESWA.2018.04.004.

[14]  S. Sabour, N. Frosst, and G. E. Hinton, "Dynamic routing between capsules," *Advances in Neural Information Processing Systems*, vol. 2017-Decem, no. Nips, pp. 3857–3867, 2017.

[15]  A. Byerly, T. Kalganova, and I. Dear, "No routing needed between capsules," *Neurocomputing*, vol. 463, pp. 545–553, Nov. 2021, doi: 10.1016/J.NEUCOM.2021.08.064.

[16]  L. Zheng *et al.*, "Spatio-temporal wind speed prediction of multiple wind farms using capsule network," *Renewable Energy*, vol. 175, pp. 718–730, 2021, doi: 10.1016/j.renene.2021.05.023.

[17]  R. Huang, J. Li, S. Wang, G. Li, and W. Li, "A Robust Weight-Shared Capsule Network for Intelligent Machinery Fault Diagnosis," *IEEE Transactions on*

*Industrial Informatics*, vol. 16, no. 10, pp. 6466–6475, 2020, doi: 10.1109/TII.2020.2964117.

[18] B. Han, "Network and Markov Transition Field / Gramian Angular Field," 2021.

[19] Z. Xu, X. Shen, Y. Wong, and M. S. Kankanhalli, "Unsupervised Motion Representation Learning with Capsule Autoencoders," no. NeurIPS, 2021, [Online]. Available: http://arxiv.org/abs/2110.00529

[20] M. Moore, "What is Industry 4.0? Everything you need to know," 2019. https://www.techradar.com/news/what-is-industry-40-everything-you-need-to-know (accessed Nov. 25, 2020).

[21] M. Jafari, "Optimal redundant sensor configuration for accuracy increasing in space inertial navigation system," *Aerospace Science and Technology*, vol. 47, pp. 467–472, 2015, doi: 10.1016/j.ast.2015.09.017.

[22] H. S. Kim, S. K. Park, Y. Kim, and C. G. Park, *Hybrid fault detection and isolation method for UAV inertial sensor redundancy management system*, vol. 16, no. 1. IFAC, 2005. doi: 10.3182/20050703-6-cz-1902.02005.

[23] S. Zhang and Z. Q. Lang, "SCADA-data-based wind turbine fault detection: A dynamic model sensor method," *Control Engineering Practice*, vol. 102, no. 322430, p. 104546, 2020, doi: 10.1016/j.conengprac.2020.104546.

[24] A. Zimek and P. Filzmoser, "There and back again: Outlier detection between statistical reasoning and data mining algorithms," *Wiley Interdisciplinary Reviews: Data Mining and Knowledge Discovery*, vol. 8, no. 6, pp. 1–26, 2018, doi: 10.1002/widm.1280.

[25] P. Gangsar and R. Tiwari, "Signal based condition monitoring techniques for fault detection and diagnosis of induction motors: A state-of-the-art review," *Mechanical Systems and Signal Processing*, vol. 144, p. 106908, 2020, doi: 10.1016/j.ymssp.2020.106908.

[26] S. Lin, R. Clark, R. Birke, and S. Sch, "ANOMALY DETECTION FOR TIME SERIES USING VAE-LSTM HYBRID MODEL," *Ieee*, pp. 4322–4326, 2020.

[27] R. Sabir, D. Rosato, S. Hartmann, and C. Gühmann, "LSTM based Bearing Fault Diagnosis of Electrical Machines using Motor Current Signal," pp. 613–618, 2019, doi: 10.1109/ICMLA.2019.00113.

[28] R. Wang, Z. Feng, S. Huang, X. Fang, and J. Wang, "Research on Voltage Waveform Fault Detection of Miniature Vibration Motor Based on Improved WP-LSTM," 2020.

[29] A. Byerly, T. Kalganova, and I. Dear, "No routing needed between capsules," *Neurocomputing*, vol. 463, pp. 545–553, 2021, doi: 10.1016/j.neucom.2021.08.064.

[30] F. Deng, S. Pu, X. Chen, Y. Shi, T. Yuan, and P. Shengyan, "Hyperspectral image classification with capsule network using limited training samples," *Sensors (Switzerland)*, vol. 18, no. 9, 2018, doi: 10.3390/s18093153.

[31] O. I. Provotar, Y. M. Linder, and M. M. Veres, "Unsupervised Anomaly Detection in Time Series Using LSTM-Based Autoencoders," *2019 IEEE International Conference on Advanced Trends in Information Theory, ATIT 2019 - Proceedings*, pp. 513–517, 2019, doi: 10.1109/ATIT49449.2019.9030505.

[32] A. Byerly and T. Kalganova, "Homogeneous Vector Capsules Enable Adaptive Gradient Descent in Convolutional Neural Networks," *arXiv*, 2020.

[33] P. Afshar, A. Mohammadi, and K. N. Plataniotis, "Brain Tumor Type Classification via Capsule Networks," in *Proceedings - International Conference on Image Processing, ICIP*, 2018, pp. 3129–3133. doi: 10.1109/ICIP.2018.8451379.

[34] M. E. Paoletti *et al.*, "Capsule Networks for Hyperspectral Image Classification," *IEEE Transactions on Geoscience and Remote Sensing*, vol. 57, no. 4, pp. 2145–2160, 2019, doi: 10.1109/TGRS.2018.2871782.

[35] R. Huang, J. Li, W. Li, and L. Cui, "Deep Ensemble Capsule Network for Intelligent Compound Fault Diagnosis Using Multisensory Data," *IEEE Transactions on Instrumentation and Measurement*, vol. 69, no. 5, 2020, doi: 10.1109/TIM.2019.2958010.

[36] S. R. R. Fahim, S. K. Sarker, S. M. Muyeen, M. R. I. Sheikh, S. K. Das, and M. G. Simoes, "A Robust Self-Attentive Capsule Network for Fault Diagnosis of Series-Compensated Transmission Line," *IEEE Transactions on Power Delivery*, vol. 8977, no. c, 2021, doi: 10.1109/TPWRD.2021.3049861.

[37] U. B. Parikh, B. Das, and R. P. Maheshwari, "Combined wavelet-SVM technique for fault zone detection in a series compensated transmission line," *IEEE Transactions on Power Delivery*, vol. 23, no. 4, pp. 1789–1794, 2008, doi: 10.1109/TPWRD.2008.919395.

[38] Z. Wang and T. Oates, "Encoding time series as images for visual inspection and classification using tiled convolutional neural networks," *AAAI Workshop - Technical Report*, vol. WS-15-14, pp. 40–46, 2015.

[39] Y. Liang, B. Li, and B. Jiao, "A deep learning method for motor fault diagnosis based on a capsule network with gate-structure dilated convolutions," *Neural Computing and Applications*, vol. 33, doi: 10.1007/s00521-020-04999-0.

[40] C. Zhao, X. Huang, Y. Li, and S. Li, "A Novel Cap-LSTM Model for Remaining Useful Life Prediction," *IEEE Sensors Journal*, vol. 21, no. 20, pp. 23498–23509, 2021, doi: 10.1109/JSEN.2021.3109623.

[41] T. Han, R. Ma, and J. Zheng, "Combination bidirectional long short-term memory and capsule network for rotating machinery fault diagnosis," *Measurement*, vol. 176, p. 109208, May 2021, doi: 10.1016/J.MEASUREMENT.2021.109208.

[42] "Bearing Data Center | Case School of Engineering | Case Western Reserve University." https://engineering.case.edu/bearingdatacenter/ (accessed Jan. 31, 2022).

[43] L. Deng, X. Wang, F. Jiang, and R. Doss, "EEG-based emotion recognition via capsule network with channel-wise attention and LSTM models," *CCF Transactions on Pervasive Computing and Interaction*, vol. 3, no. 4, pp. 425–435, Dec. 2021, doi: 10.1007/S42486-021-00078-Y/FIGURES/9.

[44] G. Hinton and R. Salakhutdinov, "Reducing the dimensionality of data with neural networks," *Science*, vol. 313, no. 5786, pp. 504–507, 2006, doi: 10.1126/science.1127647.

[45] I. D. Katser and V. O. Kozitsin, "Skoltech Anomaly Benchmark (SKAB)," *Kaggle*, 2020. https://github.com/waico/SKAB

[46] A. Lavin and S. Ahmad, "Evaluating real-time anomaly detection algorithms - The numenta anomaly benchmark," *Proceedings - 2015 IEEE 14th International Conference on Machine Learning and Applications, ICMLA 2015*, pp. 38–44, 2016, doi: 10.1109/ICMLA.2015.141.

[47] F. Tony Liu, K. Ming Ting, and Z.-H. Zhou, "Isolation Forest ICDM08," *Icdm*, 2008, [Online]. Available: https://cs.nju.edu.cn/zhouzh/zhouzh.files/publication/icdm08b.pdf%0Ahttps://cs.nju.edu.cn/zhouzh/zhouzh.files/publication/icdm08b.pdf?q=isolation-forest

[48] K. C. Gross, R. M. Singer, S. W. Wegerich, J. P. Herzog, R. VanAlstine, and F. Bockhorst, "Application of a Model-based Fault Detection System to Nuclear Plant Signals," *International conference on intelligent systems applications to power systems*, no. October 2015, p. 6, 1997, [Online]. Available: http://www.osti.gov/bridge/product.biblio.jsp?osti_id=481606

[49] C. Zhang *et al.*, "A deep neural network for unsupervised anomaly detection and diagnosis in multivariate time series data," *33rd AAAI Conference on Artificial Intelligence, AAAI 2019, 31st Innovative Applications of Artificial Intelligence Conference, IAAI 2019 and the 9th AAAI Symposium on Educational Advances in Artificial Intelligence, EAAI 2019*, pp. 1409–1416, 2019, doi: 10.1609/aaai.v33i01.33011409.

[50] P. Vijay, "Timeseries anomaly detection using an Autoencoder," *Keras*, 2020. https://keras.io/examples/timeseries/timeseries_anomaly_detection/

[51] P. Filonov, A. Lavrentyev, and A. Vorontsov, "Multivariate Industrial Time Series with Cyber-Attack Simulation: Fault Detection Using an LSTM-based Predictive Data Model," pp. 1–8, 2016, [Online]. Available: http://arxiv.org/abs/1612.06676

[52] F. Chollet, "Building Autocoders in Keras," *The Keras Blog*, 2016. https://blog.keras.io/building-autoencoders-in-keras.html

[53] S. R. Bowman, L. Vilnis, O. Vinyals, A. M. Dai, R. Jozefowicz, and S. Bengio, "Generating sentences from a continuous space," *CoNLL 2016 - 20th SIGNLL Conference on Computational Natural Language Learning, Proceedings*, pp. 10–21, 2016, doi: 10.18653/v1/k16-1002.

[54] J. Chen, S. Sathe, C. Aggarwal, and D. Turaga, "Outlier detection with autoencoder ensembles," *Proceedings of the 17th SIAM International Conference on Data Mining, SDM 2017*, pp. 90–98, 2017, doi: 10.1137/1.9781611974973.11.

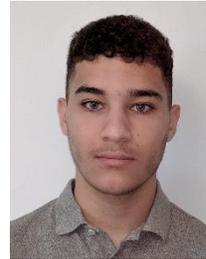

**Ayman Elhalwagy** received the BEng (Hons) degree in electronic and computer Engineering from Brunel University London, Uxbridge in 2021 and is pursuing the Ph.D. degree in Electronic and Computer Engineering with Brunel University London, Uxbridge.

His research interests include Neural Networks, Anomaly Detection and Fault Classification as well as intelligent systems.

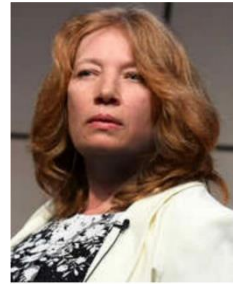

**Tatiana Kalganova** received the B.Sc. (Hons.) and Ph.D. degrees.,

She is currently a Reader in intelligent systems and the ECE Postgraduate Research Director in ECE with Brunel University London, Uxbridge, U.K. She has over 20 years of experience in design and implementation of applied intelligent systems.